\documentclass{article}

\PassOptionsToPackage{numbers, sort, compress}{natbib}

     \usepackage[preprint]{neurips_2023}

\usepackage{graphicx}
\usepackage[utf8]{inputenc} 
\usepackage[T1]{fontenc}    
\usepackage{hyperref}       
\usepackage{url}            
\usepackage{booktabs}       
\usepackage{amsfonts}       
\usepackage{nicefrac}       
\usepackage{microtype}      
\usepackage{xcolor}         

\usepackage{tikz} 
\usetikzlibrary{shapes,decorations,arrows,calc,arrows.meta,fit,positioning}
\tikzset{
    -Latex,auto,node distance =1 cm and 1 cm,semithick,
    state/.style ={ellipse, draw, minimum width = 0.7 cm},
    point/.style = {circle, draw, inner sep=0.04cm,fill,node contents={}},
    bidirected/.style={Latex-Latex,dashed},
    el/.style = {inner sep=2pt, align=left, sloped}
}

\usepackage{amssymb}
\usepackage{amsthm}
\usepackage{amsmath}
\usepackage{mathtools}

\usepackage[capitalize,noabbrev]{cleveref}

\usepackage{algorithm}
\usepackage{algpseudocode}

\usepackage{subcaption}

\theoremstyle{plain}
\newtheorem{theorem}{Theorem}[section]
\newtheorem{proposition}[theorem]{Proposition}

\theoremstyle{definition}
\newtheorem{definition}[theorem]{Definition}

\theoremstyle{remark}
\newtheorem{remark}[theorem]{Remark}

\def \bC{\mathbf{C}}
\def \bc{\mathbf{c}}
\def \bv{\mathbf{v}}
\def \bV{\mathbf{V}}

\def \bU{\mathbf{U}}
\def \bx{\mathbf{x}}
\def \bX{\mathbf{X}}
\def \bS{\mathbf{S}}

\def \bG{\mathbf{G}}
\def \bPA{\mathbf{PA}}

\def \doI{\text{do}(I)}
\def \doIC{\text{do}(I, C)}
\def \doIJ{\text{do}(I, J)}
\def \doJ{\text{do}(J)}
\def \doK{\text{do}(K)}

\def \Ehat{\hat{\mathbb{E}}}

\def \cG{\mathcal{G}}
\def \cM{\mathcal{M}}

\def \cE{\mathcal{E}}

\def \fhat{\hat{f}}
\def \Ef{\cE{(\hat{f})}}
\def \Efj{\cE_j{(\hat{f})}}

\title{Causal Dependence Plots}

\author{%
  Joshua~R.~Loftus \\
    Department of Statistics\\
    London School of Economics\\
    London, England, UK \\
  \texttt{j.r.loftus@lse.ac.uk} \\
  \And
    Lucius~E.~J.~Bynum \\
    Center for Data Science\\
    New York University\\
    New York, NY, USA \\
    \texttt{lucius@nyu.edu} \\
    \And
Sakina~Hansen \\
    Department of Statistics\\
    London School of Economics\\
    London, England, UK \\
    \texttt{s.a.hansen1@lse.ac.uk} \\
}

\begin{document}

\maketitle

\begin{abstract}
    Explaining artificial intelligence
    or machine learning models
    is increasingly important. To use such data-driven systems wisely we must understand how they interact with the world, including how they depend causally on data inputs. In this work we develop Causal Dependence Plots (CDPs) to visualize how one variable---an outcome---depends on changes in another variable---a predictor---\emph{along with any consequent causal changes in other predictor variables}. Crucially, CDPs differ from standard methods based on holding other predictors constant or assuming they are independent. CDPs make use of an auxiliary causal model because causal conclusions require causal assumptions. With simulations and real data experiments, we show CDPs can be combined in a modular way with methods for causal learning or sensitivity analysis. Since people often think causally about input-output dependence, CDPs can be powerful tools in the xAI or interpretable machine learning toolkit and contribute to applications like scientific machine learning and algorithmic fairness.
  
\end{abstract}

\section{Introduction} 
\label{sec:intro}

This paper develops Causal Dependence Plots (CDPs) to visualize causal relationships between predictor variables and an outcome variable. The idea is general, but we are motivated by explaining or interpreting AI or machine learning models \citep{guidotti_survey_2018, carvalho_machine_2019, gunning_darpas_2021, molnar_imlbook}. For simplicity we consider supervised learning where a set of features is used to predict an outcome, i.e. regression or classification.
We also focus on the model-agnostic or "black-box" explanation setting, where the interpreter can query the model but not access its internal structure. Interpretation methods in this setting are more broadly applicable for distributed research, but are also functionally limited to observing how the model responds to variation in the inputs. Our general approach has extensions beyond this initial application.

Visualizations and simple explanations that focus on one input variable at a time can be powerful tools for human understanding. Two popular visualization methods, the Partial Dependence Plot (PDP) of \citet{friedman2001greedy} and Individual Conditional Expectation (ICE) plot from \citet{goldstein2015peeking}, show how model output depends on one input variable. However, just as with the interpretation of linear regression model coefficients, the relationships revealed by focusing on one predictor at a time can be misleading. When varying one input variable, \emph{we must make some choice about what values to use for the other inputs}. PDPs treat other predictors as independent of the one being plotted, so they can correctly capture the model's dependence on each variable if predictors are independent and the model is additive \citep{hastie1986gam}. In general, when explaining the black-box's dependence on one input, our choice of how to handle other model inputs may break or respect the existing statistical or causal dependencies between predictors. This leads us to the following:

\textbf{Problem statement.} 
If there are causal relationships between predictors in the real world, but our visualization, interpretation, or explanation method does not respect them, then the resulting model explanation may be irrelevant or misleading for real world purposes \cite{moraffah2020causal, shin2021effects}. For example, such explanations could lead to incorrect decisions about regulating or aligning algorithmic systems, sub-optimal allocations of resources based on model predictions, a breakdown between human feedback and reinforcement learning systems, or other forms of harm. In the context of scientific machine learning---where explanations can be used to generate hypotheses for follow-up investigation---a flawed interpretation may direct us toward spurious hypotheses. For these reasons, we may care about the causal validity of model explanations.

\paragraph{Our proposal.} At a high level, we use an auxiliary Explanatory Causal Model (ECM) to interpret or explain a given machine learning model. For each input predictor that we wish to explain, we use the ECM to determine how other inputs vary when that predictor is manipulated, rather than treating them as independent or fixed. We call the resulting plots Causal Dependence Plots or CDPs.

Causal models can be designed based on the desired explanation, specified with prior domain knowledge, and/or potentially learned and estimated from data.

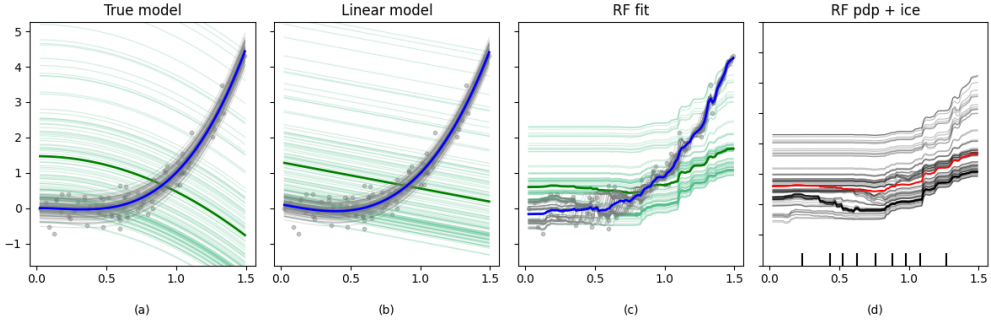
\begin{figure*}
\centering
\begin{subfigure}[t]{0.95\textwidth}
    \includegraphics[width=\textwidth]{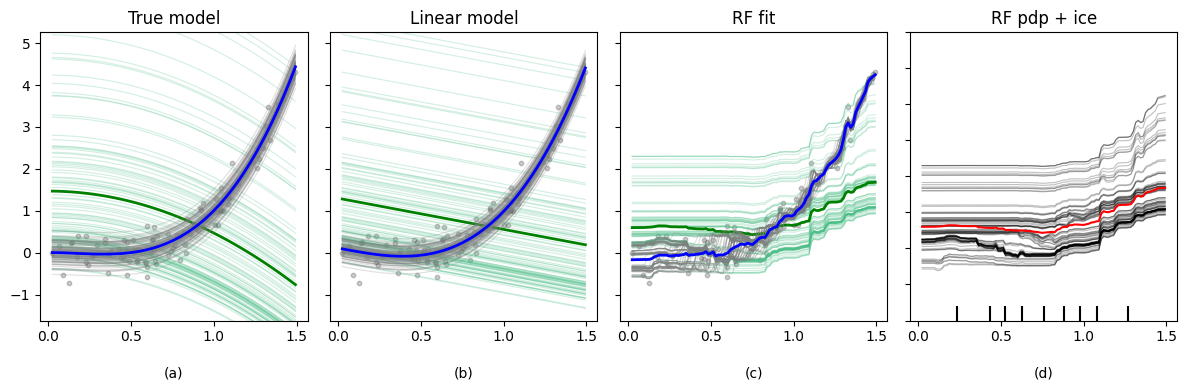}
    \label{fig:pdp_comparison}
\end{subfigure}
\begin{subfigure}[t]{0.95\textwidth}
\centering
\begin{tikzpicture}
    \node (1) [state] at (0,0) {$P$};
    \node (2) [state, right = of 1] {$F$};
    \node (3) [right = of 2] {$\hat S = \hat f(P, F)$};

    \path (1) edge (2);
    \path[densely dotted,line width=.8] (2) edge (3);
    \path[densely dotted,line width=.8, bend right=25] (1) edge (3);
\end{tikzpicture}
\end{subfigure}
\caption{{Motivating example}. Causal Dependence Plots (top) and structural graph of the Explanatory Causal Model (bottom) for the motivating example. In the top row, panel (a) shows the relationships of the ECM. Total Dependence is shown in blue and Natural Direct Dependence in green. Counterfactual curves for individual points are shown as thin, light lines, with aggregates displayed as thick, dark lines. Panels (b-c) show our model explanation plots for a linear model and random forest model, respectively. Panel (d) shows a standard Partial Dependence Plot (red) with Individual Conditional Expectation curves, for comparison to our NDDP (green).}
\label{fig:salary_example}
\end{figure*}

\paragraph{Motivating example.}
Before turning to the full details, we illustrate the idea with a simple example. Consider a mediation model for parental income $P$, school funding $F$, and graduates' average starting salary $S$, with structural equations $P \sim \mathcal{U}[0,1.5]$; $F = 2P^3 + \mathcal{N}(0,0.2^2)$; and $S = F - P^2 + \mathcal{N}(0,0.2^2)$. The corresponding directed acyclic graph (DAG) is shown in the bottom row of Figure~\ref{fig:salary_example}, with data plotted in the left panel of the top row, and the remaining panels show visual explanations of supervised models that predict $\hat S = \hat f(P, F)$.

Several important takeaways stand out from this display. First, the differences between blue and green curves show \emph{there can be qualitative differences between direct (or partial) dependence and total dependence}, a fact which is highly consequential when we consider how causal interventions may change outcomes. Second, \emph{explanations of models can be qualitatively different from the underlying causal relationships}. For example, even a flexible model like the random forest in panel (c) shows a direct dependence of $\hat S$ on $P$ that is increasing when the direct dependence of $S$ on $P$ in the true model is decreasing. As another example, panel (b) shows that {the total dependence of a linear model on a predictor can be non-linear} because the mediator $F$ depends non-linearly on $P$. Finally, we see that \emph{our framework of using causal models to produce explanation plots includes, as special cases, some existing model explanation plots like ICE and PDPs}. We revisit this point later.

\section{Methodology}
\label{sec:methods}

We give background notation and definitions in
Sections \ref{sec:learning}-\ref{sec:scm} and our methods in Sections \ref{sec:plots}-\ref{sec:uncertainty}.

\subsection{Explaining Supervised Learning Models}
\label{sec:learning}

For concreteness, we consider the supervised learning setting with a set of predictor features $\bX$ and outcome variable $Y$. We wish to explain or interpret a predictive model represented by a function $\fhat$ which is typically estimated or learned using empirical risk minimization (ERM)
$$
\fhat = \arg \min_{h \in \mathcal{H}} \sum_{i=1}^n \ell \left(h(\bx_i), y_i \right)
$$
with some loss function $\ell$, pre-specified function class $\mathcal{H}$, and an independent and identically distributed training sample $\{ (y_i, \bx_i) : i = 1, \ldots, n \}$ with feature vectors $\bx_i^T \in \mathbb{R}^p$. In \ref{sec:mediation} we focus on a specific interpretive task known as mediation analysis, again for concreteness and because it is a highly applicable example. In that section we partition the predictor variables into subsets so that $X$ and $M$ both notate predictors, $M$ being a mediator.

\paragraph{Model-agnostic explanations.}

Sometimes, for practical reasons, an explanation method does not have access to the internal mathematical structure of $\fhat$. In this model-agnostic setting we can generate explanations based on input-output dependence by providing \emph{synthetic inputs}
$$\tilde \bX \mapsto \fhat \left(\tilde{\bX}\right),$$
recording the associated predictions, and then summarizing these in some way. We denote a model-agnostic explanation generated this way as $\cE(\fhat; \tilde \bX)$ or $\Ef$ for shorthand, but note that all such explanations, including our proposed method, depend on both $\fhat$ and the synthetic input. We often target one feature at a time for ease of interpretation and creating low-dimensional visualizations. We denote an explanation targeting feature $j$ as $\Efj$, and in this case the synthetic input is
$$
\tilde{\bx}_j := (x_1, \ldots, \tilde x_j, \ldots, x_p) \mapsto \fhat \left(\tilde{\bx}_j\right).
$$
If feature $j$ is numeric, then it is typically varied along a grid in its domain. In most existing model-agnostic explanation methods, the values for other features are held fixed at observed values in a dataset used to generate the explanation. Note that this explanatory dataset may not be the same as the training data for $\fhat$. We emphasize this by notating the data used to generate an explanation as $\{ (y_i', \bx_i') : i = 1, \ldots, m \}$, where $y_i'$ may not be supplied or required depending on the type of explanation. Bar graphs can be used when the explanatory feature is categorical. \\

\begin{definition}[Individual Conditional Expectation (ICE) Plot]
\label{def:ice}
We obtain a separate curve for each observation in the explanatory dataset $x_i'$, $i = 1, \ldots, m$, by plotting the map
$$
\tilde x_j \mapsto \fhat \left(\tilde{\bx}_{ij}'\right), \text{ where }
\tilde{\bx}_{ij}' := (x_{i1}', \ldots, \tilde x_j, \ldots, x_{ip}').
$$
An ICE plot for feature $j$ displays all $m$ of these curves. \\
\end{definition}

\begin{definition}[Partial Dependence Plot (PDP)]
\label{def:pdp}
The PDP for feature $j$ can be obtained from the ICE plot for $x_j$ by computing the empirical mean over the explanatory data at each point in the grid, that is
$
\tilde x_j \mapsto \frac{1}{m} \sum_{i=1}^m \fhat \left(\tilde{\bx}_{ij}'\right).
$
\end{definition}

There are a variety of other model-agnostic explanation methods, see for example \cite{molnar_imlbook}. But since our current proposal is a visualization, these are the main alternative methods for comparison.

\paragraph{Fundamental problem of univariate explanations.} To create a model-agnostic explanation of model dependence on a single feature, like a plot with $x_j$ on the horizontal axis and $\fhat$ on the vertical axis, \emph{we must decide what to do with the other features} when synthetically varying $\tilde x_j$. Nearly all existing explanation methods use the same approach as the PDP and ICE plots: they \emph{hold other features fixed} at values in an auxiliary, explanatory dataset. This may be unrealistic if, for example, other features depend on $x_j$ causally. And it may not be mathematically defined or allow any interpretation if features are mutually constitutive, e.g. population, GDP, and GDP per capita.

\subsection{Structural Causal Models}
\label{sec:scm}

Our notational conventions and definitions below are influenced by \cite{pearl2000models, peters2017elements, bynum2023counterfactuals}. Letting $\bU$ be a set of exogeneous noise variables, $\bV$ a set of $p = |\bV|$ observable variables, and $\bG$ a set of functions such that for each $j \in 1, \ldots, p$ we have $V_j = g_j(\bPA_j, U_j)$, where $\bPA_j \subseteq \bV$ and $U_j \subseteq \bU$ are the observable and exogeneous parents, respectively, of variable $V_j$. Let the directed acyclic graph (DAG) $\cG$ have vertices given by variables and, for each $V_j \in \bV$ and each of the parent variables in $\bPA_j$ and $U_j$, a directed edge oriented from the parents to $V_j$. This graph can be useful for explanations by showing visually which variables are inputs in each function in $\bG$. \\

\begin{definition}[Structural Causal Model (SCM)]
\label{def:scm}
A (probabilistic) SCM $\cM$ is a tuple $\left< \bU, \bV, \bG, P_{\bU} \right>$ where $P_{\bU}$ is the joint distribution of the exogeneous variables. This distribution and the functions $\bG$ determine the joint distribution $P^\cM$ over all the variables in $\cM$. Finally, causality in this model is represented by additional assumptions that $\cM$ admits the modeling of interventions and/or counterfactuals as defined below. \\
\end{definition}

\begin{definition}[Interventions]
\label{def:intervention}
For the SCM $\cM$, an intervention $I$ produces a modified SCM denoted $\cM^{\doI}$ which may have different structural equations $\bG^I$. Correspondingly, some variables may have different parent sets, so the DAG representation $\cG^{\doI}$ may also change. We denote the new, interventional distribution as $P^{\cM; \doI}$. A simple class of interventions involves intervening on one variable, e.g.
$$I = \text{do}\left(V_j \coloneqq \tilde{g}(\tilde{\bPA}_j, \tilde{U}_j)\right),$$
which changes how $V_j$ and all variables on directed paths from $V_j$ in $\cG$ are generated.
An even simpler sub-class of these are the atomic interventions setting one variable $V_j$ to one constant value $v$, which we denote $I_{j,v} := \text{do}(V_j = v)$. Note that in this case $V_j$ has no parents in the graph $\cG^{\doI}$; the source of the intervention itself is outside the world of the model.
\end{definition}

Interventions are useful for modeling changes to a data generating process (DGP), for example, experiments that control a particular variable to see how its value changes other variables, or policy changes aimed at altering or removing existing causal relationships.
In addition to generating new observations as a DGP, an SCM can also be used to model counterfactual values for observations that have already been determined. A counterfactual distribution is an interventional distribution defined over a specific dataset with information or constraints given by some of the observed values in that data, as we now describe. Here we slightly abuse notation by letting boldface represent the dataset, e.g., $\bV$ are the observed variables for all observations in a previously generated dataset. \\

\begin{definition}[Counterfactuals]
\label{def:counterfactual}
For variable $V_j$ with observed values of its parents $\bPA_j = v$, we may hold some or all of $v$ fixed and vary $U_j := u$, passing these through $g_j(v, u)$ (or $\tilde{g}_j(\tilde{v}, u)$ if we also do an intervention that changes any of the values in $\bPA_j$). The counterfactuals $V_j(\tilde{v}, u)$ are values $V_j$ would have taken if any of its observed and/or exogeneous parents had taken the different values $(\tilde{v}, u)$. With intervention $I$, to define the counterfactual distribution $P^{\cM \mid \bV = \bv; \doI}$, we use the posterior or conditional (depending on our probability model approach) distribution $P_{\bU \mid \bV = \bv}$ to model uncertainty about $\bU$ while computing counterfactual values of any variables for a previously generated observation in the modified SCM $\cM^{\doI}$.
\end{definition}

\subsection{Univariate Causal Explanations}
\label{sec:causal}

\begin{figure}[ht]
\centering
\begin{subfigure}[t]{0.45\textwidth}
\centering
\begin{tikzpicture}
    \node[state] (x) at (0,0) {$\cG_\bX$};
    \node (E) [right = of x] {$\Ef$}; 

    \path[densely dotted,line width=.8] (x) edge (E);
\end{tikzpicture}
\caption{}
\label{fig:explanationG}
\end{subfigure}\hfill
\begin{subfigure}[t]{0.45\textwidth}
\centering
\begin{tikzpicture}
    \node[state] (x) at (0,0) {$X$};
    \node[state] (m) [below = of x] {$M$};
    \node[state] (y) [right = of m] {$Y$};
    \node (E) [right = of y] {$\Ef$}; 

    \path (x) edge (m);
    \path (x) edge (y);
    \path (m) edge (y);
    \path[densely dotted,line width=.8] (x) edge (E);
    \path[densely dotted,line width=.8,bend right=25] (m) edge (E);
\end{tikzpicture}
\caption{}
\label{fig:mediation}
\end{subfigure}
\caption{A structural causal model for predictors $\cG_\bX$ is used to produce an explanation $\Ef$ of the predictive model $\hat f$. In the mediation example (b), predictor $X$ causes $Y$ directly and also through mediator $M$. Solid arrows represent causal relationships in the data generation process, and dotted arrows show the formal dependence of the model explanation on predictors.}
\label{fig:explanation}
\end{figure}
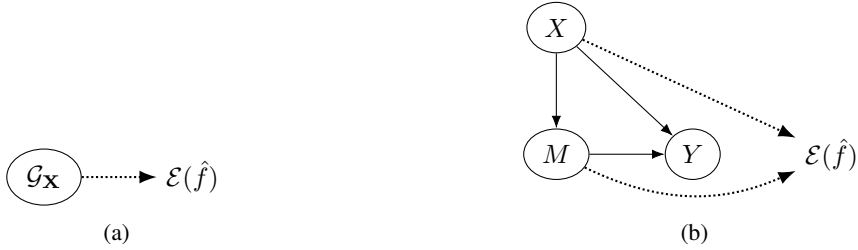

Our proposed solution to the fundamental problem highlighted for univariate explanations is to use an auxiliary ECM $\cM_j$ and let this causal model determine how other features vary as functions of $x_j$. We denote these explanations as $\cE_j(\fhat; \cM_j)$ or $\Efj$ if the context is clear. In the deterministic or noiseless case, suppose we know functions $g_{kj}$ such that $x_k = g_{kj}(x_j)$, with $g_{jj}$ the identity. In this case the model $\cM_j$ tells us
$$
x_j \mapsto g(x_j) =: \left(g_{1j}(x_j), \ldots, x_j, \ldots, g_{pj}(x_j)\right)
$$
is a curve in $\mathbb{R}^p$ parameterized by $x_j$, and we generate the explanation $\Efj$ using
$$
\tilde x_j \mapsto \fhat \left( g(\tilde x_j) \right).
$$
Next, a few more definitions will let us extend this strategy to more general, non-deterministic causal models.
We propose generating various kinds of causal explanations of a supervised learning model $\fhat$  (potentially a black-box) using an auxiliary ECM $\cM_\bX$ that captures causal relationships among the predictor variables. With $\cG_\bX$ the associated DAG, we represent this graphically in Figure~\ref{fig:explanation}, where the arrow from the subgraph $\cG_\bX$ to the explanation $\Ef$ is dotted to distinguish it from arrows among the features. Different explanations correspond to various causal operations, such as interventions, performed in $\cM_\bX$.

\begin{remark}
\label{rem:counterfactual}
Note that if the desired causal explanation uses counterfactuals, then we likely obtain the observed values from an auxiliary explanatory dataset. But since an SCM can generate data, we may also use it to generate the initial observed values and then re-use these when computing counterfactuals for the explanation. That is, we may generate the auxiliary explanatory dataset $\bV$ from $P^{\cM}$ and then, with an intervention $I$, generate counterfactuals from $P^{\cM \mid \bV = \bv; \doI}$.
\end{remark}

The generality and flexibility of SCMs allow us to pose many different interpretive questions, and since SCMs can generate synthetic data, we can use them to compute many different kinds of explanations. In the next two sections, we focus on using plots as explanations and describe several canonical types of interpretive questions. 

\subsection{Causal Dependence Plots}
\label{sec:plots}

One key motivation for this work is to define various causal analogues of the PDP and ICE plots, which we are now prepared to do. For the following definitions, we assume predictor variables $\bX \in \Omega_{\bX}$, an outcome of interest $Y \in \Omega_Y$, and a black-box function $\fhat(x): \Omega_{\bX} \rightarrow \Omega_Y$ with outputs that we may also denote $\hat{Y}$. A structural causal model $\cM_{\bX}$ is either assumed or learned from data. Importantly, $\cM_{\bX}$ specifies the causal relationships \emph{only for the predictors $\bX$} and need not involve the outcome $Y$.

\paragraph{Individual counterfactuals and expected effects.}
We use the shorthand $\fhat(P^{\cM})$, where $\fhat$ takes a distribution $P^{\cM}$ as its argument, to denote using data from that distribution as the input to the black-box function $\fhat$. For each of our plots, we show both a set of individual counterfactual curves $\fhat(P^{\cM \mid \bV = \bv; \doI})$ and their empirical average over the explanatory dataset

\begin{equation}
    \Ehat \left[ \fhat(P^{\cM \mid \bV = \bv; \doI}) \right].  
\end{equation}

For each type of causal explanation with a given \emph{Named Effect}, we define the \emph{Individual Counterfactual Name Effect} curves as the set of counterfactual curves $\fhat(P^{\cM})$ for each individual in the explanatory dataset, the \emph{Name Effect Function} as their expectation, and the \emph{Name Dependence Plot} as a plot displaying all of these curves (or rather, an empirical estimate in the case of the Effect Function). 

Before defining different types of CDPs, we first introduce a useful abstraction. Generating causal explanations involves performing abduction, action, and prediction with a structural causal model that is augmented to include the predictions we wish to explain.

\begin{definition}[Explanatory Causal Model (ECM)]
\label{def:ECM}
An ECM $\cM'$ augments the original SCM $\cM_{\bX}$ by including the predicted outcome $\hat{Y}$ as an additional variable with $\fhat$ as its structural equation. We can then, for example, compute $\fhat(P^{\cM_{\bX} \mid \bV = \bv; \doI})$ using the ECM as $P_{\hat{Y}}^{\cM' \mid \bV = \bv; \doI}$.
We describe the construction of an ECM in Algorithm~\ref{alg:explanatory_scm}.
\end{definition}

We use this process in Algorithms~\ref{alg:tdp_algorithm} through \ref{alg:nidp_algorithm} to compute each of the effects we now describe. We begin with perhaps the most straightforward and important causal effect. \\

\begin{definition}[Total Dependence Plot (TDP)]
\label{def:TE}
For an intervention $I$, the Individual Counterfactual Total Effect (ICTE) curves
\begin{equation}
\textsf{TE}(I) = \fhat(P^{\cM_{\bX} \mid \bX = \bx; \doI})
\end{equation}
show the total effect of intervention $I$ on black-box output for each individual observation in the explanatory dataset. The empirical average of these over the explanatory data is a (Monte Carlo) estimate of the Total Effect Function (TEF), and a plot showing the ICTE and TEF is a Total Dependence Plot (TDP).
We calculate the TDP following Algorithm~\ref{alg:tdp_algorithm}. \\
\end{definition}

\begin{remark}
In the remaining definitions, we give notation only for the individual counterfactual curves and leave the other objects implicitly defined.
\end{remark}


We often wish to decompose how much of the total effect of $X$ on $\hat Y$ (or $Y$) is attributable to different possible pathways between the variables. This involves understanding several causal quantities in addition to the total effect described above, which we now define. \\

\begin{definition}[Partially Controlled Dependence Plot (PCDP)] Consider intervention $I$ affecting some subset of variables in DAG $\cG_{\bX}$ and atomic intervention $C$ that holds constant any other subset of variables not intervened upon by $I$. The Individual Counterfactual Partially Controlled Effect curves
\begin{equation}
\textsf{PCE}(I, C) = \fhat(P^{\cM_{\bX} \mid \bX = \bx; \doIC})
\end{equation}
represent the effect of intervention $I$ on black-box output $\hat{Y}$ while other variables are set (via intervention) to constant values.
We compute the PCDP via Algorithm~\ref{alg:pcdp_algorithm}. \\
\end{definition}

\begin{definition}[Natural Direct Dependence Plot (NDDP)]\label{def:nde}
Consider atomic intervention $I$ and a corresponding intervention $J$ that intervenes on all children of any nodes that are intervened upon by intervention $I$ and sets them to their observed values in dataset $\bx$. For example, if $I=\text{do}(A=a, B=b)$, then intervention $J$ will set all children of the variables $A$ and $B$ to their observed values in $\bx$. We then define the Individual Counterfactual Natural Direct Effect curves as
\begin{equation}
\textsf{NDE}(I) = \fhat(P^{\cM_{\bX} \mid \bX = \bx; \doIJ}).
\end{equation}
This quantity represents the effect of intervention $I$ on black-box output $\hat{Y}$ while all variables not intervened upon are set to their `natural,' i.e., pre-intervention values in the explanatory dataset. 
Algorithm~\ref{alg:nddp_algorithm} demonstrates how to compute the NDDP.
\end{definition}

From this construction of NDDP, we see by comparing it to Definition~\ref{def:pdp} that it is equivalent to the PDP, confirming what we observed in Figure~\ref{fig:salary_example}. \\

\begin{proposition}{Partial dependence plots show natural direct effects.}
\label{prop:pdp}
When generating plots using explanatory data $D \sim \cM_\bX$, the ICE plot curves and Individual Counterfactual Natural Dependence curves---and hence also the PDP and NDDP plots---are identical. \\
\end{proposition}

\begin{definition}[Natural Indirect Dependence Plot (NIDP)]
Consider atomic intervention $I$ and a corresponding intervention $K$ that removes from DAG $\cG_{\bX}$ all outgoing edges from any of the nodes intervened upon by intervention $I$ and sets those nodes to their observed values in the explanatory dataset. For example, if $I=\text{do}(A=a, B=b)$, then intervention $K$ will remove all outgoing edges from $A$ and $B$ and set $A$ and $B$ to their original observed values. We then define Individual Counterfactual Natural Indirect Effect curves
\begin{equation}
\textsf{NIE}(I) = \fhat(P^{\cM^{\doI}_{\bX} \mid \bX = \bx; \doK}).
\end{equation}
Notice that intervention $I$ is performed before intervention $K$. This quantity represents the effect of intervention $I$ on black-box output $\hat{Y}$ that is due only to any indirect pathways to $\hat{Y}$.
We compute the NIDP following Algorithm~\ref{alg:nidp_algorithm}. The difference between two values of this function can be used to express the natural indirect effect as a special case.
\end{definition}

The Supplementary Material includes full descriptions of algorithms for computing all of the above plots.

\subsection{Mediation Analysis}
\label{sec:mediation}

Many applications involve a causal structure we refer to as a mediation triangle, with an example shown in Figure~\ref{fig:mediation}. In mediation analysis, we often wish to decompose how much of the total effect of $X$ on $Y$ is attributable to the pathway through $M$ and how much of it is direct. Our above definitions allow us to visualize frequently studied quantities of interest in this setting. For example, the difference between two values of the PCDP can be used to express the controlled direct effect as a special case --- specifically, with interventions $I, C$ defined as $I = \text{do}(X=x)$ and $C = \text{do}(M=m)$ in the mediation triangle. Although mediation analysis motivates CDPs and helps build intuition, we emphasize that the definitions in Section~\ref{sec:plots} can be used \emph{with any structural causal model}. See Section~\ref{sec:experiments} for other, more complex examples.

\subsection{Incorporating Uncertainty in Causal Dependence}
\label{sec:uncertainty}

\begin{figure}
    \centering
    \includegraphics[width=\textwidth]{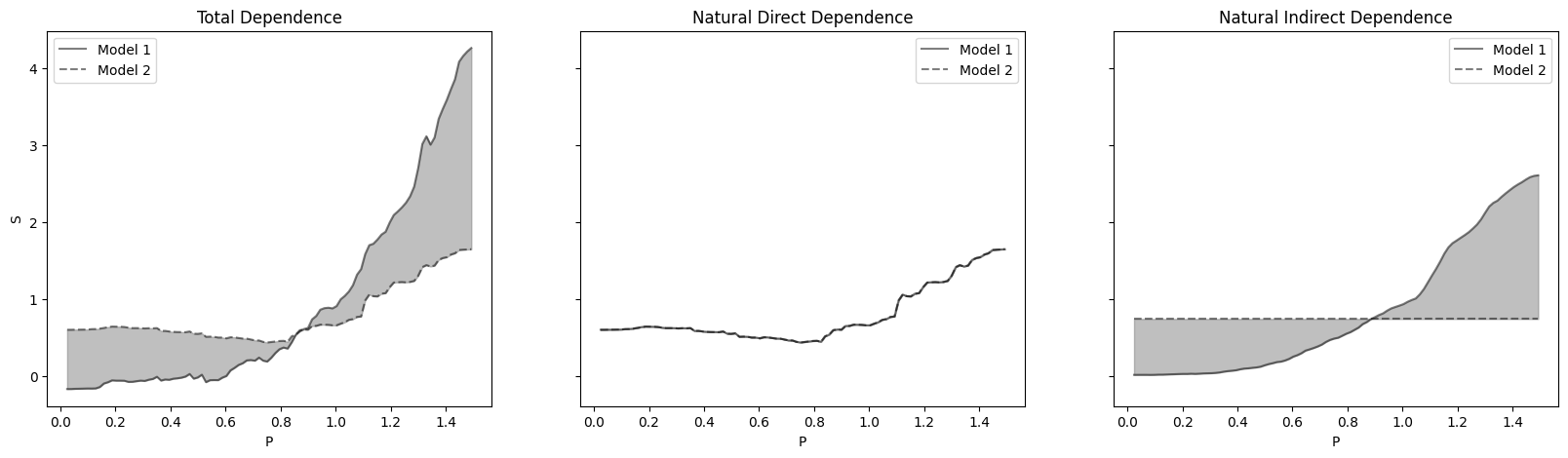}
    \caption{TDP, NDDP, and NIDP for the salary example that incorporate causal uncertainty, now visualizing a range of possible effect functions induced by two competing candidate causal models --- one with mediation and one without.}
    \label{fig:salary_uncertainty}
\end{figure}

There are various ways to incorporate uncertainty about the true causal model into CDPs. We now explore a natural first extension of the CDP that shows a \emph{range of possible effect functions} induced by a \emph{set of auxiliary ECMs}. The set of ECMs could be pre-specified or, for example, a Markov equivalence class of DAGs output by a causal structure learning algorithm.
Returning to our motivating example from Section~\ref{sec:intro}, we might question whether parental income $P$ impacts school funding $F$, considering instead an SCM without mediation: $P \rightarrow S \leftarrow F$. Figure~\ref{fig:salary_uncertainty} shows a range of possible effect functions interpolating between this ECM without the indirect effect and the original ECM in Section~\ref{sec:intro}, for each of the TDP, NDDP, and NIDP. In this we have assumed the same structural equations for the edges that are common to both models. These plots show a range for how predictions might depend on one predictor $P$ when we are unsure how the other predictor depends on $P$. In the Supplementary Material we show an example with real data where we use candidate ECMs discovered by the PC algorithm.

In this example, we have shown how incorporating multiple causal models into CDPs allows us to directly visualize our uncertainty about the underlying causal model and its impact on the effect we expect a predictor $X$ to have on black-box output $\hat{Y}$. More broadly, this process allows us to characterize how $X$ will impact $Y$ or $\hat{Y}$ {under multiple conditions}, enabling a versatile sensitivity analysis.

\section{Experiments}
\label{sec:experiments}

We now illustrate several kinds of causal explanations our method can produce.

\begin{figure*}
    \centering
    \includegraphics[width=\textwidth]{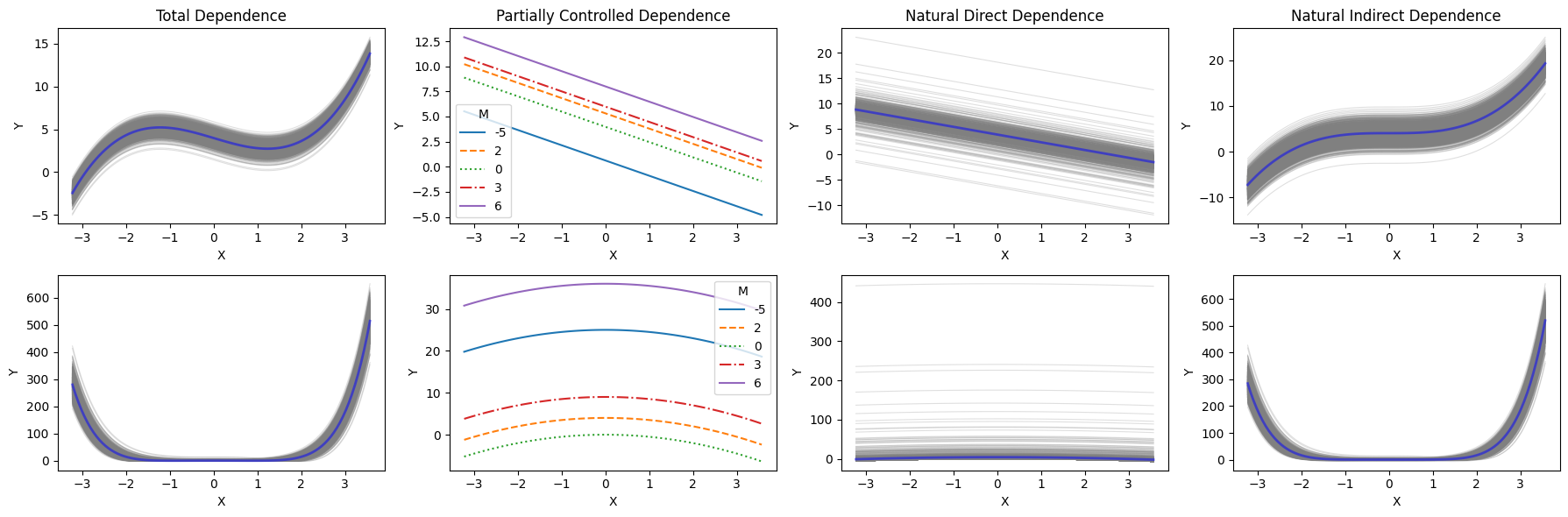}
    \caption{CDPs for the simulation example in Section~\ref{sec:experiments}, shown for both a `correct' black-box model (bottom row) and `incorrect' black-box model (top row).}
    \label{fig:mediator_grid}
\end{figure*}

\paragraph{Simulations.} Consider the non-linear mediation example, governed by the following DGP. 
\begin{equation*} \label{eu_eqn}
    X \sim \mathcal{N}(0,1), \quad
    M = \frac{1}{2} X^3 + \mathcal{N}(0,1), \quad
    Y = M^2 - \frac{1}{2}X^2 + \mathcal{N}(0,1).
\end{equation*}
We use this DGP to fit two different black-box models:  one model that assumes the correct functional form (i.e., the relationship for $Y$ shown in the DGP above), and an `incorrect' model that predicts $Y$ via linear regression. We use the Python \texttt{DoWhy} package \citep{dowhypaper, dowhy_gcm} in our experiments to sample counterfactual data in the construction of our plots. Figure~\ref{fig:mediator_grid} shows the CDPs for each of these models using the black-box training data as the explanatory data. 
We can glean a couple insights from Figure~\ref{fig:mediator_grid}. First, the CDPs are all sensitive to whether the functional form assumptions of the black-box model fit the ground truth data generating process. The second is that the different effects on the outcome $Y$ that we may want to investigate will show up visually across the different plots. For example, the TDP for the incorrect model shows dependence on $X$ that looks cubic, while the true relationship is quadratic and sextuplic. By looking at the NIDP and NDDP, we can see that for the incorrect model the cubic relationship is due to an indirect effect through mediator $M$. The PCDP for the correct vs. incorrect model shows us directly how the quadratic term --- the controlled direct effect on $Y$ --- is either present or not in the black-box model.

\paragraph{Real data with causal discovery.} The Breast Cancer Wisconsin (Original) dataset \citep{street1993nuclear} is a publicly available dataset often used to test algorithms on medical data. The dataset contains 9 ordinal variables, which represent attributes of the cells within a breast mass: Clump Thickness, Uniformity of Cell Size, Uniformity of Cell Shape, Marginal Adhesion, Single Epithelial Cell Size, Bare Nuclei, Bland Chromatin, Normal Nucleoli and Mitoses. The outcome variable is the class of the breast tumor, benign or malignant. 

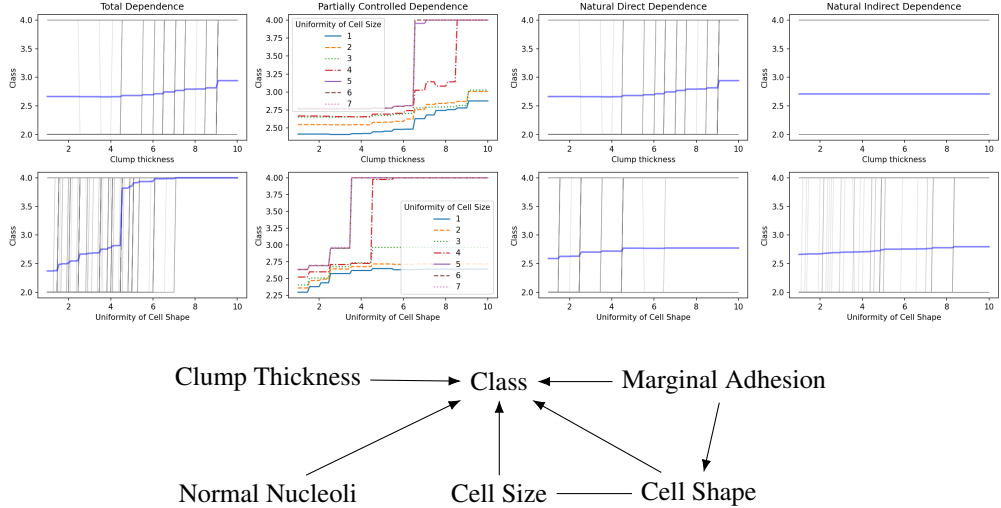
\begin{figure}[ht]
\centering
\begin{subfigure}[t]{0.95\textwidth}
    \centering
    \includegraphics[width=\textwidth]{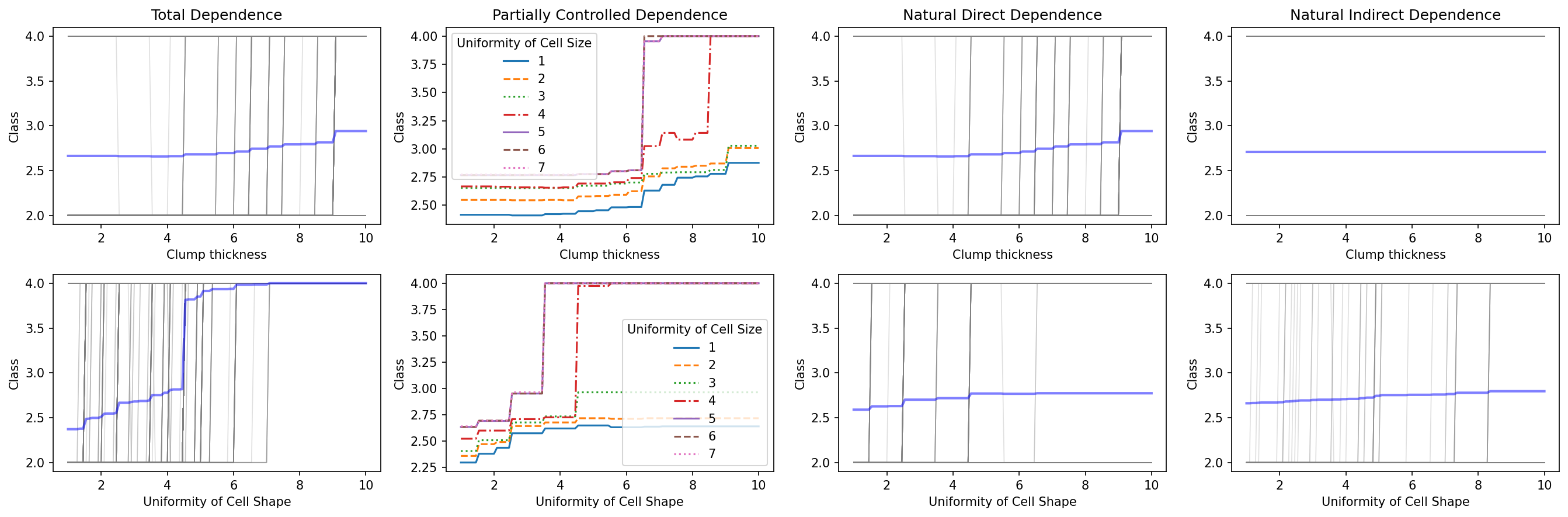}
    \label{fig:bcancer_grid}
\end{subfigure}
\begin{subfigure}[t]{0.95\textwidth}
\centering
\begin{tikzpicture}
    \node (1) at (0,0) {Cell Size};
    \node (3) [right = of 1] {Cell Shape};
    \node (4) [above = of 1] {Class};
    \node (5) [right = of 4] {Marginal Adhesion};
    \node (6) [left = of 1] {Normal Nucleoli};
    \node (2) [above = of 6] {Clump Thickness};    
    \path (1) edge (4);
    \path (2) edge (4);
    \draw[-] (1) -- (3);
    \path (5) edge (3);
    \path (3) edge (4);
    \path (6) edge (4);
    \path (5) edge (4);
\end{tikzpicture}
\label{sec:breast cancer}
\end{subfigure}
    \caption{Breast cancer data example. CDPs for a random forest classifier and predictors Clump Thickness (first row) and Uniformity of Cell Shape (second row). Structural graph $\cG_{B}$ for the ECM learned by the PC algorithm (last row). The outcome Class is binary: 2 for benign, 4 for malignant.}
    \label{fig:breast_cancer}
\end{figure}

We use a causal structural learning algorithm, specifically the PC algorithm \cite{spirtes2000causation} implemented in Julia \texttt{CausalInference}, to learn a DAG for this dataset, on a smaller subset of predictor variables for simplicity. Figure~\ref{fig:breast_cancer} shows the resulting DAG and CDPs for a random forest model to classify the Class variable.

This shows CDPs can be combined with other causal methods like structural learning algorithms. The PC algorithm output had an undirected edge between Cell Size and Cell Shape, so we investigate sensitivity to the different graph structures consistent with this output in the Supplementary Material.

\section{Applications, Extensions, Related Work}
\label{sec:applications}

{Explanations under covariate shift:} Often a model that has already been trained will be used for predictions on data that may not follow the training DGP. If knowledge about causal relationships in the shifted covariate distribution is available, we can leverage that to choose an ECM, and CDPs can visualize how the model will behave under covariate shift.

{Causal semi-supervised learning:} Given knowledge of causal structure among predictors and a supervised learning model to predict an outcome, our method can be used to attempt inference of causal relationships from the predictors to the outcome, similar in spirit to \cite{scholkopf2012causal} and \cite{zhao_causal_2021}.

{Auditing for fairness or other desiderata:} Previous work has applied causal methodology to fairness \citep{kilbertus2017avoiding, kusner_counterfactual_2017, russell_when_2017, nabi2018fair, loftus_causal_2018, zhang2018fairness, chiappa2019path, kusner_making_2019, yang_causal_2021, bynum_disaggregated_2021, makhlouf2020survey}, recourse \citep{ustun_actionable_2019, poyiadzi_face_2020, karimi_towards_2022}, and other desiderata. CDPs can be used to probe a black-box for such properties.

{Scientific theory development:} Large and complex machine learning models may be fit to massive datasets where underlying structure is largely unknown. In such settings, relatively simple ECMs can be used to formulate hypotheses relating multiple predictors and plot various causal dependencies as a way of generating new hypotheses or checking the assumed ones for plausibility.

{Uncertainty and sensitivity analysis:} Future work can develop methods for visualizing uncertainty. For example, sensitivity analysis based on conformal prediction \citep{jin2023sensitivity, yin2022conformal, chernozhukov2021exact, lei2021conformal}. If there is uncertainty between several candidate ECMs, our method in Section~\ref{sec:uncertainty} or one analogous to that in \cite{russell_when_2017} could produce composite CDPs.

Related work: Recent work motivated by recourse \citep{karimi_survey_2021} uses contrastive or counterfactual explanations \citep{stepin_survey_2021}. Some of this work is not based on causality despite using the term ``counterfactual,'' but some does focus on causal dependence \cite{sani2020explaining}. \citet{blobaum2017estimation} produce causal explanations by identifying the predictor with the largest total effect, which is most applicable when assuming linearity. \citet{zhao_causal_2021} investigated causal interpretations of PDP. That paper aimed to use such plots for causal inference about the underlying DGP rather than as explanations of the black-box, and showed that when the DGP satisfies the backdoor criterion \citep{pearl1993adjustment} then a PDP visualizes the total effect (TE) of a predictor. \citet{cox2018modernizing} observed an association between partial dependence plots and NDE, an equivalence we formally establish in Proposition~\ref{prop:pdp}, to our knowledge the first such result. \citet{lazzari2022predicting} proposed weighting observations when computing PDPs, which could potentially be used for confounder adjustment. Our unified framework shows how disparate model interpretation plots originally proposed without causal motivation or justification can be related to causal interpretations. Aside from the coincidental cases of PDPs and M-plots \citep{molnar_imlbook} with NDE and TE, we are not aware of any previous work providing explanations or interpretive plots for the other kinds of causal explanations that fit in our framework. 

\section{Discussion}
\label{sec:conclusion}

\paragraph{Limitations.}

Causal modeling in general involves some limitations that we do not repeat here, but see for example \cite{greenland2015limitations}. Model-agnostic explanation methods are also subject to limitations \citep{molnar_imlbook, altmann2020limitations}, a few of which we will highlight because we believe they are important to keep in mind when using our method.

Mismatch between the black-box and the true DGP: if the predictive model fails to fit the DGP, then practically any model explanation will also fail if our interpretive goal is to learn about the DGP \citep{zhao_causal_2021}. This issue is particularly troubling in the model-agnostic setting where we cannot conduct model diagnostics and probe the internals of the black-box. 

Mismatch between the explanatory SCM and true/target DGP: CDPs may be misleading if the true DGP differs in important ways from the ECM. However, standard PDPs and similar explanation methods also require auxiliary explanatory data, and that data may also differ from the target DGP. So this is not much of an additional limitation specific to our method.

Availability of the ECM: if we consider models as tools and are not concerned about whether there is a ``true'' causal model for the DGP, we still need to choose which tools to use when producing an explanation. In this sense full knowledge of an SCM can be a strong assumption. However, in Sections \ref{sec:uncertainty} and \ref{sec:applications} we discussed some ways this can be improved. In general, \emph{if a causal explanation is desired or necessary, then we cannot avoid making causal assumptions}.

\paragraph{Conclusion.}
In this paper we proposed Causal Dependence Plots, a method that uses a given (potentially learned) explanatory causal model to create various plots with causal interpretations. This allows us to use the powerful language of structural causal models to pose and answer a variety of causally meaningful questions. Our framework unites previously disparate, non-causally motivated interpretive tools like partial dependence plots, and reveals some new kinds of causal interpretations we have not seen previously explored in the literature. Additional future work in this direction could explore other relatively small canonical causal structures for useful applications, or interface with other kinds of models, for example extending to non-tabular data by applying causal representation learning. Relating explanation methods to Pearl's ladder of causation \cite{pearl2018book}, most previous interpretable machine learning and explainable AI methods---like PDPs---concern associations and hence are confined to the first rung of the ladder. With CDPs we ascend the ladder, creating model interpretations intended to change the world. While interpretability provided the initial motivation for CDPs, we believe plotting such causal relationships will be useful in other settings as well.

\bibliographystyle{plainnat}
\bibliography{references.bib}

\begin{thebibliography}{49}
\providecommand{\natexlab}[1]{#1}
\providecommand{\url}[1]{\texttt{#1}}
\expandafter\ifx\csname urlstyle\endcsname\relax
  \providecommand{\doi}[1]{doi: #1}\else
  \providecommand{\doi}{doi: \begingroup \urlstyle{rm}\Url}\fi

\bibitem[Altmann et~al.(2020)Altmann, Bodensteiner, Dankers, Dassen, Fritz,
  Gruber, et~al.]{altmann2020limitations}
T~Altmann, J~Bodensteiner, C~Dankers, T~Dassen, N~Fritz, S~Gruber, et~al.
\newblock Limitations of interpretable machine learning methods.
\newblock 2020.

\bibitem[Bl{\"o}baum and Shimizu(2017)]{blobaum2017estimation}
Patrick Bl{\"o}baum and Shohei Shimizu.
\newblock Estimation of interventional effects of features on prediction.
\newblock In \emph{2017 IEEE 27th International Workshop on Machine Learning
  for Signal Processing (MLSP)}, pages 1--6. IEEE, 2017.

\bibitem[Bl{\"o}baum et~al.(2022)Bl{\"o}baum, G{\"o}tz, Budhathoki, Mastakouri,
  and Janzing]{dowhy_gcm}
Patrick Bl{\"o}baum, Peter G{\"o}tz, Kailash Budhathoki, Atalanti~A.
  Mastakouri, and Dominik Janzing.
\newblock Dowhy-gcm: An extension of dowhy for causal inference in graphical
  causal models.
\newblock \emph{arXiv preprint arXiv:2206.06821}, 2022.

\bibitem[Bynum et~al.(2021)Bynum, Loftus, and
  Stoyanovich]{bynum_disaggregated_2021}
Lucius Bynum, Joshua Loftus, and Julia Stoyanovich.
\newblock Disaggregated {Interventions} to {Reduce} {Inequality}.
\newblock In \emph{Equity and {Access} in {Algorithms}, {Mechanisms}, and
  {Optimization}}, pages 1--13. Association for Computing Machinery, New York,
  NY, USA, October 2021.
\newblock ISBN 978-1-4503-8553-4.
\newblock URL \url{https://doi.org/10.1145/3465416.3483286}.

\bibitem[Bynum et~al.(2023)Bynum, Loftus, and
  Stoyanovich]{bynum2023counterfactuals}
Lucius Bynum, Joshua Loftus, and Julia Stoyanovich.
\newblock Counterfactuals for the future.
\newblock In \emph{Proceedings of the AAAI Conference on Artificial
  Intelligence}, 2023.

\bibitem[Carvalho et~al.(2019)Carvalho, Pereira, and
  Cardoso]{carvalho_machine_2019}
Diogo~V. Carvalho, Eduardo~M. Pereira, and Jaime~S. Cardoso.
\newblock Machine {Learning} {Interpretability}: {A} {Survey} on {Methods} and
  {Metrics}.
\newblock \emph{Electronics}, 8\penalty0 (8):\penalty0 832, August 2019.
\newblock ISSN 2079-9292.
\newblock \doi{10.3390/electronics8080832}.
\newblock URL \url{https://www.mdpi.com/2079-9292/8/8/832}.
\newblock Number: 8 Publisher: Multidisciplinary Digital Publishing Institute.

\bibitem[Chernozhukov et~al.(2021)Chernozhukov, W{\"u}thrich, and
  Zhu]{chernozhukov2021exact}
Victor Chernozhukov, Kaspar W{\"u}thrich, and Yinchu Zhu.
\newblock An exact and robust conformal inference method for counterfactual and
  synthetic controls.
\newblock \emph{Journal of the American Statistical Association}, 116\penalty0
  (536):\penalty0 1849--1864, 2021.

\bibitem[Chiappa(2019)]{chiappa2019path}
Silvia Chiappa.
\newblock Path-specific counterfactual fairness.
\newblock In \emph{Proceedings of the AAAI Conference on Artificial
  Intelligence}, volume~33, pages 7801--7808, 2019.

\bibitem[Cox~Jr(2018)]{cox2018modernizing}
Louis~Anthony Cox~Jr.
\newblock Modernizing the bradford hill criteria for assessing causal
  relationships in observational data.
\newblock \emph{Critical reviews in toxicology}, 48\penalty0 (8):\penalty0
  682--712, 2018.

\bibitem[Friedman(2001)]{friedman2001greedy}
Jerome~H Friedman.
\newblock Greedy function approximation: a gradient boosting machine.
\newblock \emph{Annals of statistics}, pages 1189--1232, 2001.

\bibitem[Goldstein et~al.(2015)Goldstein, Kapelner, Bleich, and
  Pitkin]{goldstein2015peeking}
Alex Goldstein, Adam Kapelner, Justin Bleich, and Emil Pitkin.
\newblock Peeking inside the black box: Visualizing statistical learning with
  plots of individual conditional expectation.
\newblock \emph{journal of Computational and Graphical Statistics}, 24\penalty0
  (1):\penalty0 44--65, 2015.

\bibitem[Greenland and Mansournia(2015)]{greenland2015limitations}
Sander Greenland and Mohammad~Ali Mansournia.
\newblock Limitations of individual causal models, causal graphs, and
  ignorability assumptions, as illustrated by random confounding and design
  unfaithfulness.
\newblock \emph{European journal of epidemiology}, 30:\penalty0 1101--1110,
  2015.

\bibitem[Guidotti et~al.(2018)Guidotti, Monreale, Ruggieri, Turini, Giannotti,
  and Pedreschi]{guidotti_survey_2018}
Riccardo Guidotti, Anna Monreale, Salvatore Ruggieri, Franco Turini, Fosca
  Giannotti, and Dino Pedreschi.
\newblock A {Survey} of {Methods} for {Explaining} {Black} {Box} {Models}.
\newblock \emph{ACM Computing Surveys}, 51\penalty0 (5):\penalty0 93:1--93:42,
  August 2018.
\newblock ISSN 0360-0300.
\newblock \doi{10.1145/3236009}.
\newblock URL \url{https://doi.org/10.1145/3236009}.

\bibitem[Gunning et~al.(2021)Gunning, Vorm, Wang, and
  Turek]{gunning_darpas_2021}
David Gunning, Eric Vorm, Jennifer~Yunyan Wang, and Matt Turek.
\newblock {DARPA}'s explainable {AI} ({XAI}) program: {A} retrospective.
\newblock \emph{Applied AI Letters}, 2\penalty0 (4):\penalty0 e61, 2021.
\newblock ISSN 2689-5595.
\newblock \doi{10.1002/ail2.61}.
\newblock URL \url{https://onlinelibrary.wiley.com/doi/abs/10.1002/ail2.61}.
\newblock \_eprint: https://onlinelibrary.wiley.com/doi/pdf/10.1002/ail2.61.

\bibitem[Hastie and Tibshirani(1986)]{hastie1986gam}
Trevor Hastie and Robert Tibshirani.
\newblock {Generalized Additive Models}.
\newblock \emph{Statistical Science}, 1\penalty0 (3):\penalty0 297 -- 310,
  1986.
\newblock \doi{10.1214/ss/1177013604}.
\newblock URL \url{https://doi.org/10.1214/ss/1177013604}.

\bibitem[Jin et~al.(2023)Jin, Ren, and Cand{\`e}s]{jin2023sensitivity}
Ying Jin, Zhimei Ren, and Emmanuel~J Cand{\`e}s.
\newblock Sensitivity analysis of individual treatment effects: A robust
  conformal inference approach.
\newblock \emph{Proceedings of the National Academy of Sciences}, 120\penalty0
  (6):\penalty0 e2214889120, 2023.

\bibitem[Kalainathan et~al.(2019)Kalainathan, Goudet, and
  Dutta]{Kalainathan2019CausalDT}
Diviyan Kalainathan, Olivier Goudet, and Ritik Dutta.
\newblock Causal discovery toolbox: Uncovering causal relationships in python.
\newblock \emph{J. Mach. Learn. Res.}, 21:\penalty0 37:1--37:5, 2019.

\bibitem[Karimi et~al.(2021)Karimi, Barthe, Schölkopf, and
  Valera]{karimi_survey_2021}
Amir-Hossein Karimi, Gilles Barthe, Bernhard Schölkopf, and Isabel Valera.
\newblock A survey of algorithmic recourse: definitions, formulations,
  solutions, and prospects.
\newblock \emph{arXiv:2010.04050 [cs, stat]}, March 2021.
\newblock URL \url{http://arxiv.org/abs/2010.04050}.
\newblock arXiv: 2010.04050.

\bibitem[Karimi et~al.(2022)Karimi, von Kügelgen, Schölkopf, and
  Valera]{karimi_towards_2022}
Amir-Hossein Karimi, Julius von Kügelgen, Bernhard Schölkopf, and Isabel
  Valera.
\newblock Towards {Causal} {Algorithmic} {Recourse}.
\newblock In Andreas Holzinger, Randy Goebel, Ruth Fong, Taesup Moon,
  Klaus-Robert Müller, and Wojciech Samek, editors, \emph{{xxAI} - {Beyond}
  {Explainable} {AI}: {International} {Workshop}, {Held} in {Conjunction} with
  {ICML} 2020, {July} 18, 2020, {Vienna}, {Austria}, {Revised} and {Extended}
  {Papers}}, Lecture {Notes} in {Computer} {Science}, pages 139--166. Springer
  International Publishing, Cham, 2022.
\newblock ISBN 978-3-031-04083-2.
\newblock \doi{10.1007/978-3-031-04083-2_8}.
\newblock URL \url{https://doi.org/10.1007/978-3-031-04083-2_8}.

\bibitem[Kilbertus et~al.(2017)Kilbertus, Rojas~Carulla, Parascandolo, Hardt,
  Janzing, and Sch{\"o}lkopf]{kilbertus2017avoiding}
Niki Kilbertus, Mateo Rojas~Carulla, Giambattista Parascandolo, Moritz Hardt,
  Dominik Janzing, and Bernhard Sch{\"o}lkopf.
\newblock Avoiding discrimination through causal reasoning.
\newblock \emph{Advances in neural information processing systems}, 30, 2017.

\bibitem[Kusner et~al.(2019)Kusner, Russell, Loftus, and
  Silva]{kusner_making_2019}
Matt Kusner, Chris Russell, Joshua Loftus, and Ricardo Silva.
\newblock Making {Decisions} that {Reduce} {Discriminatory} {Impacts}.
\newblock In \emph{Proceedings of the 36th {International} {Conference} on
  {Machine} {Learning}}, pages 3591--3600. PMLR, May 2019.
\newblock URL \url{https://proceedings.mlr.press/v97/kusner19a.html}.
\newblock ISSN: 2640-3498.

\bibitem[Kusner et~al.(2017)Kusner, Loftus, Russell, and
  Silva]{kusner_counterfactual_2017}
Matt~J Kusner, Joshua Loftus, Chris Russell, and Ricardo Silva.
\newblock Counterfactual {Fairness}.
\newblock In \emph{Advances in {Neural} {Information} {Processing} {Systems}},
  volume~30. Curran Associates, Inc., 2017.
\newblock URL
  \url{https://proceedings.neurips.cc/paper/2017/hash/a486cd07e4ac3d270571622f4f316ec5-Abstract.html}.

\bibitem[Lazzari et~al.(2022)Lazzari, Alvarez, and
  Ruggieri]{lazzari2022predicting}
Matilde Lazzari, Jose~M Alvarez, and Salvatore Ruggieri.
\newblock Predicting and explaining employee turnover intention.
\newblock \emph{International Journal of Data Science and Analytics},
  14\penalty0 (3):\penalty0 279--292, 2022.

\bibitem[Lei and Cand{\`e}s(2021)]{lei2021conformal}
Lihua Lei and Emmanuel~J Cand{\`e}s.
\newblock Conformal inference of counterfactuals and individual treatment
  effects.
\newblock \emph{Journal of the Royal Statistical Society Series B}, 83\penalty0
  (5):\penalty0 911--938, 2021.

\bibitem[Loftus et~al.(2018)Loftus, Russell, Kusner, and
  Silva]{loftus_causal_2018}
Joshua~R. Loftus, Chris Russell, Matt~J. Kusner, and Ricardo Silva.
\newblock Causal {Reasoning} for {Algorithmic} {Fairness}.
\newblock \emph{arXiv:1805.05859 [cs]}, May 2018.
\newblock URL \url{http://arxiv.org/abs/1805.05859}.
\newblock arXiv: 1805.05859.

\bibitem[Makhlouf et~al.(2020)Makhlouf, Zhioua, and
  Palamidessi]{makhlouf2020survey}
Karima Makhlouf, Sami Zhioua, and Catuscia Palamidessi.
\newblock Survey on causal-based machine learning fairness notions.
\newblock \emph{arXiv preprint arXiv:2010.09553}, 2020.

\bibitem[Molnar(2022)]{molnar_imlbook}
Christoph Molnar.
\newblock \emph{Interpretable Machine Learning: A Guide for Making Black Box
  Models Explainable}.
\newblock 2022.
\newblock URL \url{https://christophm.github.io/interpretable-ml-book/}.

\bibitem[Moraffah et~al.(2020)Moraffah, Karami, Guo, Raglin, and
  Liu]{moraffah2020causal}
Raha Moraffah, Mansooreh Karami, Ruocheng Guo, Adrienne Raglin, and Huan Liu.
\newblock Causal interpretability for machine learning-problems, methods and
  evaluation.
\newblock \emph{ACM SIGKDD Explorations Newsletter}, 22\penalty0 (1):\penalty0
  18--33, 2020.

\bibitem[Nabi and Shpitser(2018)]{nabi2018fair}
Razieh Nabi and Ilya Shpitser.
\newblock Fair inference on outcomes.
\newblock In \emph{Proceedings of the AAAI Conference on Artificial
  Intelligence}, volume~32, 2018.

\bibitem[Pearl(1993)]{pearl1993adjustment}
Judea Pearl.
\newblock [bayesian analysis in expert systems]: Comment: Graphical models,
  causality and intervention.
\newblock \emph{Statistical Science}, 8\penalty0 (3):\penalty0 266--269, 1993.
\newblock ISSN 08834237.
\newblock URL \url{http://www.jstor.org/stable/2245965}.

\bibitem[Pearl and Mackenzie(2018)]{pearl2018book}
Judea Pearl and Dana Mackenzie.
\newblock \emph{The book of why: the new science of cause and effect}.
\newblock Basic books, 2018.

\bibitem[Pearl et~al.(2000)]{pearl2000models}
Judea Pearl et~al.
\newblock Models, reasoning and inference.
\newblock \emph{Cambridge, UK: CambridgeUniversityPress}, 19\penalty0 (2),
  2000.

\bibitem[Peters et~al.(2017)Peters, Janzing, and
  Sch{\"o}lkopf]{peters2017elements}
Jonas Peters, Dominik Janzing, and Bernhard Sch{\"o}lkopf.
\newblock \emph{Elements of Causal Inference: Foundations and Learning
  Algorithms}.
\newblock MIT Press, 2017.

\bibitem[Poyiadzi et~al.(2020)Poyiadzi, Sokol, Santos-Rodriguez, De~Bie, and
  Flach]{poyiadzi_face_2020}
Rafael Poyiadzi, Kacper Sokol, Raul Santos-Rodriguez, Tijl De~Bie, and Peter
  Flach.
\newblock {FACE}: {Feasible} and {Actionable} {Counterfactual} {Explanations}.
\newblock In \emph{Proceedings of the {AAAI}/{ACM} {Conference} on {AI},
  {Ethics}, and {Society}}, pages 344--350. Association for Computing
  Machinery, New York, NY, USA, February 2020.
\newblock ISBN 978-1-4503-7110-0.
\newblock URL \url{https://doi.org/10.1145/3375627.3375850}.

\bibitem[Ramsey and Andrews(2018)]{ramsey2018fask}
Joseph Ramsey and Bryan Andrews.
\newblock Fask with interventional knowledge recovers edges from the sachs
  model.
\newblock \emph{ArXiv}, abs/1805.03108, 2018.

\bibitem[Russell et~al.(2017)Russell, Kusner, Loftus, and
  Silva]{russell_when_2017}
Chris Russell, Matt~J Kusner, Joshua Loftus, and Ricardo Silva.
\newblock When {Worlds} {Collide}: {Integrating} {Different} {Counterfactual}
  {Assumptions} in {Fairness}.
\newblock In \emph{Advances in {Neural} {Information} {Processing} {Systems}},
  volume~30. Curran Associates, Inc., 2017.
\newblock URL
  \url{https://proceedings.neurips.cc/paper/2017/hash/1271a7029c9df08643b631b02cf9e116-Abstract.html}.

\bibitem[Sachs et~al.(2005)Sachs, Perez, Pe'er, Lauffenburger, and
  Nolan]{sachs2005paper}
Karen Sachs, Omar Perez, Dana Pe'er, Douglas~A. Lauffenburger, and Garry~P.
  Nolan.
\newblock Causal protein-signaling networks derived from multiparameter
  single-cell data.
\newblock \emph{Science}, 308\penalty0 (5721):\penalty0 523--529, 2005.
\newblock \doi{10.1126/science.1105809}.
\newblock URL \url{https://www.science.org/doi/abs/10.1126/science.1105809}.

\bibitem[Sani et~al.(2020)Sani, Malinsky, and Shpitser]{sani2020explaining}
Numair Sani, Daniel Malinsky, and Ilya Shpitser.
\newblock Explaining the behavior of black-box prediction algorithms with
  causal learning.
\newblock \emph{arXiv preprint arXiv:2006.02482}, 2020.

\bibitem[Sch{\"o}lkopf et~al.(2012)Sch{\"o}lkopf, Janzing, Peters, Sgouritsa,
  Zhang, and Mooij]{scholkopf2012causal}
B~Sch{\"o}lkopf, D~Janzing, J~Peters, E~Sgouritsa, K~Zhang, and J~Mooij.
\newblock On causal and anticausal learning.
\newblock In \emph{29th International Conference on Machine Learning (ICML
  2012)}, pages 1255--1262. International Machine Learning Society, 2012.

\bibitem[Sharma and Kiciman(2020)]{dowhypaper}
Amit Sharma and Emre Kiciman.
\newblock Dowhy: An end-to-end library for causal inference.
\newblock \emph{arXiv preprint arXiv:2011.04216}, 2020.

\bibitem[Shin(2021)]{shin2021effects}
Donghee Shin.
\newblock The effects of explainability and causability on perception, trust,
  and acceptance: Implications for explainable ai.
\newblock \emph{International Journal of Human-Computer Studies}, 146:\penalty0
  102551, 2021.

\bibitem[Spirtes et~al.(2000)Spirtes, Glymour, Scheines, and
  Heckerman]{spirtes2000causation}
Peter Spirtes, Clark~N Glymour, Richard Scheines, and David Heckerman.
\newblock \emph{Causation, prediction, and search}.
\newblock MIT press, 2000.

\bibitem[Stepin et~al.(2021)Stepin, Alonso, Catala, and
  Pereira-Fariña]{stepin_survey_2021}
Ilia Stepin, Jose~M. Alonso, Alejandro Catala, and Martín Pereira-Fariña.
\newblock A {Survey} of {Contrastive} and {Counterfactual} {Explanation}
  {Generation} {Methods} for {Explainable} {Artificial} {Intelligence}.
\newblock \emph{IEEE Access}, 9:\penalty0 11974--12001, 2021.
\newblock ISSN 2169-3536.
\newblock \doi{10.1109/ACCESS.2021.3051315}.
\newblock Conference Name: IEEE Access.

\bibitem[Street et~al.(1993)Street, Wolberg, and
  Mangasarian]{street1993nuclear}
W~Nick Street, William~H Wolberg, and Olvi~L Mangasarian.
\newblock Nuclear feature extraction for breast tumor diagnosis.
\newblock In \emph{Biomedical image processing and biomedical visualization},
  volume 1905, pages 861--870. SPIE, 1993.

\bibitem[Ustun et~al.(2019)Ustun, Spangher, and Liu]{ustun_actionable_2019}
Berk Ustun, Alexander Spangher, and Yang Liu.
\newblock Actionable {Recourse} in {Linear} {Classification}.
\newblock In \emph{Proceedings of the {Conference} on {Fairness},
  {Accountability}, and {Transparency}}, {FAT}* '19, pages 10--19, New York,
  NY, USA, January 2019. Association for Computing Machinery.
\newblock ISBN 978-1-4503-6125-5.
\newblock \doi{10.1145/3287560.3287566}.
\newblock URL \url{https://doi.org/10.1145/3287560.3287566}.

\bibitem[Yang et~al.(2021)Yang, Loftus, and Stoyanovich]{yang_causal_2021}
Ke~Yang, Joshua~R. Loftus, and Julia Stoyanovich.
\newblock Causal {Intersectionality} and {Fair} {Ranking}.
\newblock In Katrina Ligett and Swati Gupta, editors, \emph{2nd {Symposium} on
  {Foundations} of {Responsible} {Computing} ({FORC} 2021)}, volume 192 of
  \emph{Leibniz {International} {Proceedings} in {Informatics} ({LIPIcs})},
  pages 7:1--7:20, Dagstuhl, Germany, 2021. Schloss Dagstuhl –
  Leibniz-Zentrum für Informatik.
\newblock ISBN 978-3-95977-187-0.
\newblock \doi{10.4230/LIPIcs.FORC.2021.7}.
\newblock URL \url{https://drops.dagstuhl.de/opus/volltexte/2021/13875}.
\newblock ISSN: 1868-8969.

\bibitem[Yin et~al.(2022)Yin, Shi, Wang, and Blei]{yin2022conformal}
Mingzhang Yin, Claudia Shi, Yixin Wang, and David~M Blei.
\newblock Conformal sensitivity analysis for individual treatment effects.
\newblock \emph{Journal of the American Statistical Association}, pages 1--14,
  2022.

\bibitem[Zhang and Bareinboim(2018)]{zhang2018fairness}
Junzhe Zhang and Elias Bareinboim.
\newblock Fairness in decision-making—the causal explanation formula.
\newblock In \emph{Proceedings of the AAAI Conference on Artificial
  Intelligence}, volume~32, 2018.

\bibitem[Zhao and Hastie(2021)]{zhao_causal_2021}
Qingyuan Zhao and Trevor Hastie.
\newblock Causal interpretations of black-box models.
\newblock \emph{Journal of Business \& Economic Statistics}, 39\penalty0
  (1):\penalty0 272--281, 2021.
\newblock \doi{10.1080/07350015.2019.1624293}.
\newblock URL \url{https://doi.org/10.1080/07350015.2019.1624293}.

\end{thebibliography}

\clearpage
\begin{center}
\textbf{\large Supplemental Material: Causal Dependence Plots}
\end{center}
\setcounter{equation}{0}
\setcounter{figure}{0}
\setcounter{table}{0}
\setcounter{page}{1}
\makeatletter
\renewcommand{\theequation}{S\arabic{equation}}
\renewcommand{\thefigure}{S\arabic{figure}}

\section{Algorithms for Causal Dependence Plots}

In this section, we provide algorithms to compute each of the causal dependence plots defined in the main text.

\begin{algorithm}[H]
\caption{Explanatory Causal Model (ECM)\\Inputs: $\cM_{\bX}$ (SCM), $\hat{f}$ (black-box predictor), $\bS \subseteq \bX$ (covariates used by black-box)}\label{alg:explanatory_scm}
\begin{algorithmic}
\State Make copy $\cM'$ of SCM $\cM_{\bX}$
\State Add node for $\hat{Y}$ to causal graph $\cG$ of SCM $\cM'$
\For{$x$ in $\bS$}
    \State Add edge in $\cG$ from $x$ to $\hat{Y}$
\EndFor
\State Set structural equation for node $\hat{Y}$ to $\hat{f}$
\State Set exogenous variable $U_{\hat{Y}} \leftarrow 0$\\
\Return $\cM'$ 
\end{algorithmic}
\end{algorithm}

\begin{algorithm}[H]
\caption{Total Dependence Plot (TDP)\\Inputs: $\cM_{\bX}$ (SCM), $\hat{f}$ (black-box predictor), $D$ (explanatory dataset), $X_s$ (covariate of interest)}\label{alg:tdp_algorithm}
\begin{algorithmic}
\State Get ECM $\cM$ via Algorithm~\ref{alg:explanatory_scm}
\State Get the possible values of $X_S$ and set to $X$
\State Set $N$ to the number of observations in $D$
\State Initialize $N \times |X|$ matrix of estimates $\hat{Y}$
\For{$x$ in $X$}
    \State Define intervention $I = \text{do}(X_S = x)$
    \State Sample counterfactual dataset $D_c$ entailed by $P^{\cM \mid D; \doI}$
    \State Set $\hat{Y}[:, x] \leftarrow D_c[:, y]$ for index $y$ corresponding to node $\hat{Y}$
\EndFor
\State Plot $N$ lines $(X, \hat{Y}[i, :])$
\Comment{(Individual Counterfactuals)}
\State Plot average $(X, \sum_i{\hat{Y}}[i, :]/N)$ 
\Comment{(Causal Dependence)}
\end{algorithmic}
\end{algorithm}

\begin{algorithm}[H]
\caption{Partially Controlled Dependence Plot (PCDP)\\Inputs: $\cM_{\bX}$ (SCM), $\hat{f}$ (black-box predictor), $D$ (explanatory dataset), $X_s$ (covariate of interest), $C$ (intervention controlling other variables in $\cM_{\bX}$)}\label{alg:pcdp_algorithm}
\begin{algorithmic}
\State Get ECM $\cM$ via Algorithm~\ref{alg:explanatory_scm}
\State Get the possible values of $X_S$ and set to $X$
\State Set $N$ to the number of observations in $D$
\State Initialize $N \times |X|$ matrix of estimates $\hat{Y}$
\For{$x$ in $X$}
    \State Define intervention $I = \text{do}(X_S = x, C)$
    \State Sample counterfactual dataset $D_c$ entailed by $P^{\cM \mid D; \doI}$
    \State Set $\hat{Y}[:, x] \leftarrow D_c[:, y]$ for index $y$ corresponding to node $\hat{Y}$
\EndFor
\State Plot $N$ lines $(X, \hat{Y}[i, :])$
\Comment{(Individual Counterfactuals)}
\State Plot average $(X, \sum_i{\hat{Y}}[i, :]/N)$ 
\Comment{(Causal Dependence)}
\end{algorithmic}
\end{algorithm}

\begin{algorithm}[H]
\caption{Natural Direct Dependence Plot (NDDP)\\Inputs: $\cM_{\bX}$ (SCM), $\hat{f}$ (black-box predictor), $D$ (explanatory dataset), $X_s$ (covariate of interest)}\label{alg:nddp_algorithm}
\begin{algorithmic}
\State Get ECM $\cM$ via Algorithm~\ref{alg:explanatory_scm}
\State Get the possible values of $X_S$ and set to $X$
\State Set $N$ to the number of observations in $D$
\State Initialize $N \times |X|$ matrix of estimates $\hat{Y}$
\State Get all children of $X_S$ in $\cM$, excluding $\hat{Y}$, and store in $\bC$
\State Make copy $\cM'$ of SCM $\cM$
\For{$x$ in $\bC$}
    \State Remove all incoming edges to $x$ from $\cM'$
\EndFor
\For{$x$ in $X$}
    \For{$i$ in $N$}
        \State Get observed values of all variables in $\bC$ for unit $i$ and store in $\bc_i$
        \State Define intervention $I = \text{do}(X_S = x, \bC=\bc_i)$
        \State Sample counterfactual observation $d_c$ for unit $i$ entailed by $P^{\cM' \mid D[i]; \doI}$
        \State Set $\hat{Y}[i, x] \leftarrow d_c[y]$ for index $y$ corresponding to node $\hat{Y}$
    \EndFor
\EndFor
\State Plot $N$ lines $(X, \hat{Y}[i, :])$
\Comment{(Individual Counterfactuals)}
\State Plot average $(X, \sum_i{\hat{Y}}[i, :]/N)$ 
\Comment{(Causal Dependence)}
\end{algorithmic}
\end{algorithm}

\begin{algorithm}[H]
\caption{Natural Indirect Dependence Plot (NIDP)\\Inputs: $\cM_{\bX}$ (SCM), $\hat{f}$ (black-box predictor), $D$ (explanatory dataset), $X_s$ (covariate of interest)}\label{alg:nidp_algorithm}
\begin{algorithmic}
\State Get ECM $\cM$ via Algorithm~\ref{alg:explanatory_scm}
\State Get the possible values of $X_S$ and set to $X$
\State Set $N$ to the number of observations in $D$
\State Initialize $N \times |X|$ matrix of estimates $\hat{Y}$
\State Get all children of $X_S$ in $\cM$, excluding $\hat{Y}$, and store in $\bC$
\State Make copy $\cM'$ of SCM $\cM$
\For{$x$ in $\bC$}
    \State Remove all incoming edges to $x$ from $\cM'$
\EndFor
\State Define intervention $I = \text{do}(X_S = x)$
\For{$x$ in $X$}
    \For{$i$ in $N$}
        \State Sample counterfactual observation $d_c$ for unit $i$ entailed by $P^{\cM \mid D[i]; \doI}$
        \State Get counterfactual values of all variables in $\bC$ from observation $d_c$ and store in $\bc_i$
        \State Define intervention $J = \text{do}(X_S = x, \bC=\bc_i)$
        \State Sample counterfactual observation $d'_c$ for unit $i$ entailed by $P^{\cM' \mid D[i]; \doJ}$
        \State Set $\hat{Y}[i, x] \leftarrow d'_c[y]$ for index $y$ corresponding to node $\hat{Y}$
    \EndFor
\EndFor
\State Plot $N$ lines $(X, \hat{Y}[i, :])$
\Comment{(Individual Counterfactuals)}
\State Plot average $(X, \sum_i{\hat{Y}}[i, :]/N)$ 
\Comment{(Causal Dependence)}
\end{algorithmic}
\end{algorithm}

\section{Real data with causal discovery}

In this section, we explore the graph structures consistent with the uncertain edge in the DAG $\cG_{B}$ in the main text for the Breast Cancer Wisconsin dataset.

\begin{figure}[H]
    \centering
    \includegraphics[width=\textwidth]{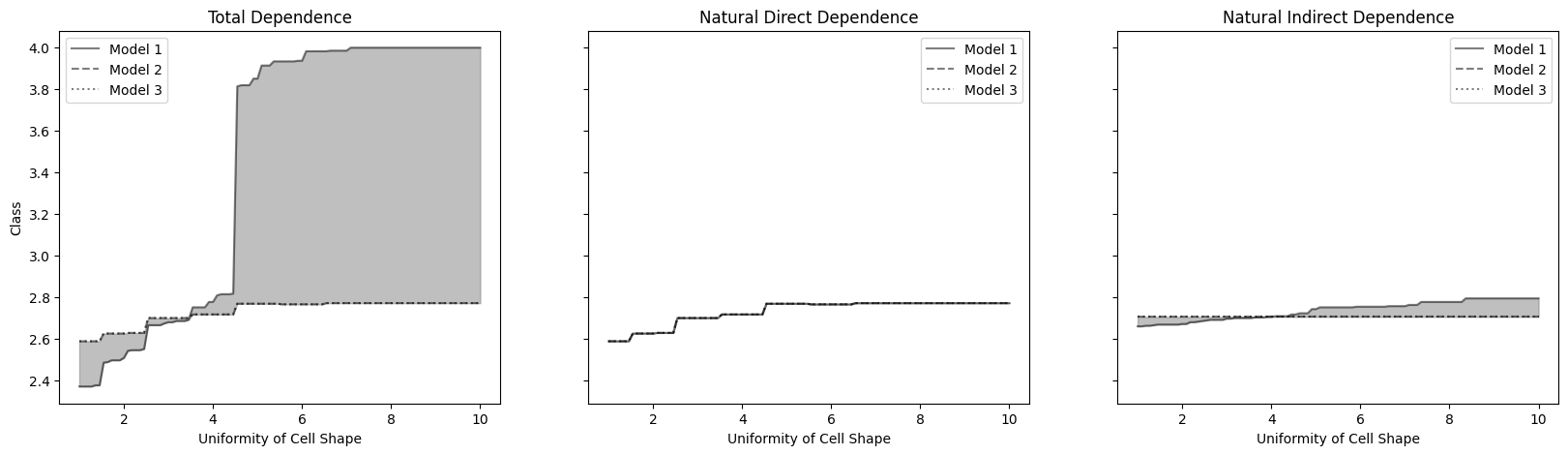}
    \caption{Total Dependence Plots, Natural Direct Dependence Plots and Natural Indirect Dependence Plots for the Breast Cancer Wisconsin dataset under three possible DAGs found by the PC algorithm: (1) $\cG_{B}$ with the edge Cell Shape $\rightarrow$ Cell Size, (2) $\cG_{B}$ with the edge Cell Size $\rightarrow$ Cell Shape, and (3) $\cG_{B}$ with no edge between Cell Size and Cell Shape.}
    \label{fig:bcancer_uncertainty}
\end{figure}

Figure~\ref{fig:bcancer_uncertainty} shows the TDP, NDDP, and NIDP for a learned additive noise model (ANM) with three different structures consistent with $\cG_{B}$: (1) an ANM with the edge Cell Shape $\rightarrow$ Cell Size, (2) an ANM with the edge Cell Size $\rightarrow$ Cell Shape, and (3) an ANM with no edge between Cell Size and Cell Shape. This figure shows that the takeaway about cell shape impacting tumor class is indeed sensitive to our choice of what to assume about the uncertain edge, particularly in the case of total dependence.

\section{Real data with domain expertise}

\begin{figure}[H]
\begin{subfigure}[c]{0.3\textwidth}
    \begin{tikzpicture}
        \node (1) at (-2,1.5) {P38};
        \node (2) at (0,1.5) {praf};
        \node (3) at (2,1.5) {pjnk};
        \node (4) at (0,3) {PKC};
        \node (5) at (2,0) {pmek};
        \node (6) at (0,0) {PKA};
        \node (7) at (2,-2) {p44/42};
        \node (8) at (-2,-2) {PIP2};
        \node (9) at (0,-2) {PIP3};
        \node (10) at (-2,0) {plcg};
        \node (11) at (0,-1) {pakts473};
    
        \path (4) edge (1);
        \path (6) edge (1);
    
        \path (4) edge (2);
        \path (6) edge (2);
    
        \path (4) edge (3);
        \path (6) edge (3);
    
        \path (8) edge (4);
        \path (10) edge (4);
        
        \path (2) edge (5);
        \path (4) edge (5);
        \path (6) edge (5);
    
        \path (5) edge (7);
        \path (6) edge (7);
    
        \path (9) edge (8);
        \path (10) edge (8);
        
        \path (9) edge (10);
    
        \path (6) edge (11);
        \path (9) edge (11);
    
    \end{tikzpicture}
    \label{fig:sachs_graph}
\end{subfigure}\hfill
\begin{subfigure}[c]{0.6\textwidth}
    \includegraphics[width=\textwidth]{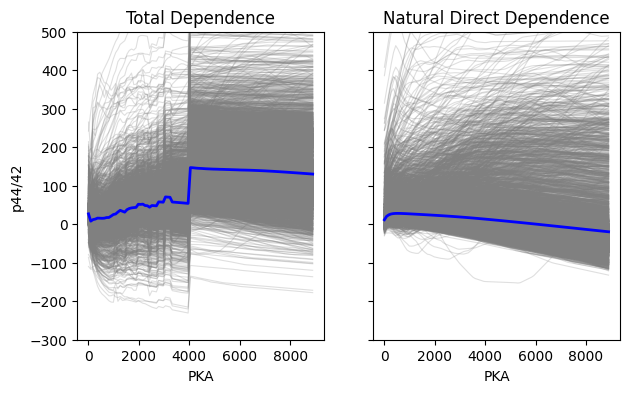}
\end{subfigure}
\caption{DAG $\cG_{S}$ for the \citet{sachs2005paper} dataset as well as a corresponding Total Dependence Plot and Natural Direct Dependence Plot for the effect of PKA on p44/42.}
\label{fig:sachs_example}
\end{figure}

As an example that makes use of domain expertise for the underlying causal model, Figure~\ref{fig:sachs_example} shows a DAG, TDP, and NDDP for the \citet{sachs2005paper} dataset, for which data and a ground-truth DAG\footnote{Following the discussion in \citet{ramsey2018fask} and follow-up ground truth DAG for the \citet{sachs2005paper} dataset in \citeauthor{ramsey2018fask}'s Figure~5, we choose the edge PIP3 $\rightarrow$ PIP2 in order to eliminate a would-be cycle, and otherwise leave the released DAG unchanged.} are publicly available in the Causal Discovery Toolbox \cite{Kalainathan2019CausalDT}.  While the actual biology of the problem is not our focus here, there are meaningful implications from Figure~\ref{fig:sachs_example}. From this model, the TDP shows a relationship between the two features that is importantly different from the NDDP. Recall that the NDDP captures the same relationship we would see from a PDP or an ICE plot. If we don't consider how other variables in the graph will change in response to changes in PKA, the trend we uncover will be practically reversed.

\end{document}